%% file: iclr2025_conference.tex
\documentclass{article} 
\usepackage{iclr2027_conference}
\usepackage{times}
\input{math_commands.tex}

\usepackage{hyperref}
\usepackage{url}

\title{FlashDiffusion: Fused Tiled Kernel Spectral Decomposition}

\author{Julio Candanedo \\
SparseTrace LLC \\
Appleton, WI 54913, USA \\
\texttt{julio@sparsetrace.ai} \\
}

\usepackage{amssymb,mathrsfs,amsthm}
\usepackage{booktabs}

\iclrfinalcopy 
\usepackage{pgfplots}
\usepackage{pgfplotstable}
\usepgfplotslibrary{groupplots}
\pgfplotsset{compat=1.18}

\begin{document}

\maketitle

\begin{abstract}
Diffusion maps, and kernel methods more generally, provide an interpretable nonlinear spectral representation basis for geometric learning. In the geometric limit, small bandwidth, these matrices tend to be high rank and thus require materializing dense Gaussian kernels requires $\mathcal{O}(N^2)$ memory. We introduce \textsc{FlashDiffusion}, a matrix-free method that evaluates dense Gaussian kernel blocks in fused GPU tiles and couples the eigensolver to an empirical $\beta$-flow that selects the finite-sample resolution scale. A continuation over sample size and bandwidth warm-starts increasingly expensive spectral solves from coarser resolutions.
\end{abstract}



\input{sections/introduction}

\input{sections/FlashDiffusion}

\input{sections/T6experiment}

\input{sections/AlanineDipeptide}

\input{sections/Conclusion}

\newpage
\bibliography{iclr2025_conference}
\bibliographystyle{iclr2025_conference}

\newpage
\appendix
\input{sections/ANN} 

\input{sections/keops_compare}

\end{document}

%% file: math_commands.tex
\usepackage{amsmath,amsfonts,bm}

\def\eqref#1{equation~\ref{#1}}

\def\1{\bm{1}}

\DeclareMathAlphabet{\mathsfit}{\encodingdefault}{\sfdefault}{m}{sl}
\SetMathAlphabet{\mathsfit}{bold}{\encodingdefault}{\sfdefault}{bx}{n}



%% file: sections/introduction.tex
\section{Introduction}\label{sec:intro}

Kernel nonparametric methods and their associated manifold-learning methods, e.g. Laplacian eigenmaps \citep{BelkinNiyogi2003_LaplacianEigenmaps}, offer an interpretable route to nonlinear representation learning: nonlinear correlations $\mathsf{K}=\exp_\odot(-\beta\mathcal{H})$ are obtained by exponentiating a matrix of linear correlations $\mathcal{H}\in\mathbb{R}^{N\times N}$, e.g. pairwise squared distances. They face two related difficulties. First, the pairwise kernel requires $\mathcal{O}(N^2)$ arithmetic and, if materialized, $\mathcal{O}(N^2)$ memory. Second, the bandwidth $\beta$ fundamentally changes the spectrum and effective rank of the kernel. Prior work has addressed the computational problem with $k$NN graphs, \cite{muja2014_scalableNearestNeighbors}, low-rank approximations and fast transforms.\footnote{Random Fourier Features \citep{rahimi2007_randomFeatures,rahimi2008_kitchenSinks}, Nyström features \citep{williams2000_nystromKernelMachines, long2019_landmarkDiffusionMaps}, and the fast Gauss transform \citep{greengard1991_fastGaussTransform,raykar2005_improvedFastGaussTransform,hertrich2024_fastKernelSummation}.} We discuss $k$NN separately in \S\ref{sec:knn}, since it sparsifies spatial interactions and hence fundamentally different than truncating spectral rank. Low-rank methods face a distinct limitation; at the bandwidths required to resolve a substantial manifold spectrum, the Gaussian kernel is generally high rank, so aggressive compression removes spectral directions the diffusion representation is intended to recover. Bandwidth selection has meanwhile been approached through median or quantile rules, locally adaptive $k$-th-neighbor bandwidths \citep{zelnikmanor2004_selfTuningSpectralClustering}, and the log--log kernel-sum criterion \citep{coifman2008_graphLaplacianTomography}; these provide useful scale estimates, but depend on the dataset's sampling and geometry and do not directly optimize finite-sample spectral resolution.

We propose that both problems are coupled, and should be treated together. With finite data, the diffusion matrix approximates an underlying continuum operator. Singer and Hein \citep{singer2006_graphToManifoldLaplacian,hein2007_graphLaplaciansRandomNeighborhoodGraphs} showed that the kernel length scale required for consistency depends on sampling density and hence on $N$. In our parameterization, increasing $\beta$ decreases the spatial kernel length scale, so greater sampling can support a larger $\beta$ and resolve finer structure. We therefore seek an empirical resolution scale $\beta^\star(N)$, selected before the leading spectrum reaches the sampling and numerical resolution floors. Two limits clarify the picture: as $\beta\to0$, the kernel approaches its broad-kernel, low-rank linear limit \citep{candanedo2025_linearizedDiffusionMap}; along a continuum-consistent path $N\to\infty$ with $\beta=\beta^\star(N)$, progressively finer manifold structure becomes resolvable and the kernel develops a correspondingly large effective rank \citep{clemente2018_laplaceBeltramiCDT}.

Recent work connects self-attention with diffusion maps \citep{candanedo2026_DAC}, suggesting a computational strategy for the quadratic pairwise operator. FlashAttention \citep{dao2022_flashAttention,zadouri2026_flashAttention4} addresses the analogous $\mathcal{O}(N^2)$ memory obstruction in attention by evaluating pairwise interactions in GPU tiles rather than materializing the full matrix. \textsc{FlashDiffusion} applies the same IO-aware principle to the Gaussian kernel: pairwise diffusion interactions are evaluated in parallel GPU tiles and used immediately as matrix-block products inside a block Lanczos eigensolver \citep{lanczos1950_iterationEigenvalueProblem,baglama2003_irbl}, avoiding explicit materialization of the $\mathcal{O}(N^2)$ kernel while retaining the dense interaction at the chosen precision.

\textsc{FlashDiffusion} remains formally $\mathcal{O}(N^2)$ in pairwise arithmetic, and each Lanczos iteration requires another quadratic kernel application. At small $\beta$ the broad Gaussian kernel has rapidly decaying spectrum, so the leading eigenspace is inexpensive to resolve. Increasing $\beta$ reveals additional spectral directions and makes the eigensolve progressively more demanding. We therefore bootstrap across sample size and bandwidth: starting from a subsample $n\ll N$, \textsc{FlashDiffusion} finds $\beta^\star(n)$ and its eigenspace at cost $\mathcal{O}(n^2)$, extends the eigenspace to a larger sample by the Nyström extension \citep{bengio2003_outOfSample}, and re-optimizes $\beta$. Repeating this constructs an $(n,\beta)$ ladder following the empirical resolution curve $\beta^\star(n)$ toward the full dataset. For nested subsamples drawn from the same underlying distribution, increased sampling density progressively resolves finer spectral structure, allowing the full-$N$ solve to begin from a well-adapted eigenspace. \textsc{FlashDiffusion} therefore couples bandwidth selection, dense kernel evaluation, and eigenspace continuation while preserving the high-rank diffusion operator in the regime where its resolved manifold structure is largest.

%% file: sections/FlashDiffusion.tex
\section{\textsc{FlashDiffusion} Algorithm}\label{sec:flashdiffusion}

\subsection{Matrix-free diffusion operator}\label{sec:matrixfree}

Given $N$ observations $\mathcal{R}\in\mathbb{R}^{N\times D}$, diffusion maps begin with the Gaussian affinity operator:
\begin{align}\label{eq:kernel}
\mathsf{K}=\exp_{\odot}\left(-\beta\, \mathcal{H}\right),
\end{align}
where $\mathcal{H}$ contains the pairwise squared Euclidean distances. Let $q=\mathsf{K}\mathbf{1}$. The Coifman--Lafon normalization \citep{coifman2006_diffusionMaps} defines $\mathsf{K}_{\alpha}=\mathrm{diag}(q)^{-\alpha}\,\mathsf{K}\,\mathrm{diag}(q)^{-\alpha}$, with degree $d=\mathsf{K}_{\alpha}\mathbf{1}$ and row-stochastic diffusion operator $\mathsf{P}=\mathrm{diag}(d)^{-1}\mathsf{K}_{\alpha}$. Although $\mathsf{P}$ is generally asymmetric, it is similar to the symmetric operator (with $s=q^{-\alpha}\odot d^{-1/2}$, powers taken element-wise):
\begin{align}\label{eq:symmetric}
\mathsf{M} &=\mathrm{diag}(d)^{1/2}\,\mathsf{P}\,\mathrm{diag}(d)^{-1/2} =\mathrm{diag}(s)\,\mathsf{K}\,\mathrm{diag}(s).
\end{align}
Thus $\mathsf{P}$ and $\mathsf{M}$ share the same eigenvalues, and the diffusion spectrum can be obtained with a symmetric block Lanczos eigensolver. Note that each fresh $\beta$ pays an entry fee before any Lanczos iteration can run: the normalization vectors are themselves kernel applications, $q=\mathsf{K}\mathbf{1}$ and, for $\alpha\neq0$, $d=\mathrm{diag}(q)^{-\alpha}\,\mathsf{K}\,\mathrm{diag}(q)^{-\alpha}$, two full $\mathcal{O}(N^2)$ passes that must complete before $s$ exists. This is a structural departure from attention, which normalizes each softmax row on-the-fly within a single pass, \cite{milakov2018_onlineSoftmax}. The resulting operator is symmetrizable, eq.~\ref{eq:symmetric}, unlike self-attention and this purchase is what licenses the symmetric eigensolver and its guarantees. Once $s$ is computed, each block Lanczos iteration reduces to the matrix-block operation, for a block $V\in\mathbb{R}^{N\times b}$:
\begin{align}\label{eq:matmat}
\mathsf{M}V=\mathrm{diag}(s)\,\mathsf{K}\left(\mathrm{diag}(s)V\right).
\end{align}
Normalization passes and block Lanczos iterations therefore share a single computational primitive: the Gaussian application $W\mapsto\mathsf{K}W$, with $W=\mathbf{1}$, $q^{-\alpha}$, or $\mathrm{diag}(s)V$. This primitive can be evaluated directly from the observations. With mean-centered coordinates $\bar{\mathcal{R}}=\mathcal{R}-\mathbf{1}\mu^\top$, define $\mathcal{G}=\bar{\mathcal{R}}\bar{\mathcal{R}}^\top$ and $r=\mathrm{diag}(\mathcal{G})$. Translation preserves pairwise distances (and centering bounds $r$, taming cancellation in the distance form), hence:
\begin{align}\label{eq:rbf_gram}
\mathsf{K}=\exp_{\odot}\left(-\beta\left(r\mathbf{1}^\top+\mathbf{1}r^\top-2\mathcal{G}\right)\right).
\end{align}
The kernel is thus a function of Gram blocks of the observations, and never needs to exist as a stored matrix: its tiled evaluation is the subject of \S\ref{sec:rbfxgemm}.

\subsection{XGEMM: Tiled Kernel Operator}\label{sec:rbfxgemm}

FlashAttention reduces memory traffic by constructing pairwise interactions in fast on-chip memory rather than materializing the full quadratic matrix in HBM \citep{dao2022_flashAttention}; \textsc{FlashDiffusion} applies the same IO-aware principle to the Gaussian kernel application in eq.~\ref{eq:matmat}. Unlike the attention operator, the diffusion operator is scaled on both sides, $\mathrm{diag}(s)\,\mathsf{K}\,\mathrm{diag}(s)$, and $s$ couples the Lanczos block to global functions of the kernel, and hence no online construction exists. The costs are nonetheless contained, the $q$ and $d$ passes are themselves kernel applications (\S\ref{sec:matrixfree}) and reuse the identical tile schedule with $V$ replaced by $\mathbf{1}$ and $q^{-\alpha}$; and within an operator pass the right scaling is the precomputed block $\mathrm{diag}(s)V$ while the left scaling is a row rescale fused into the accumulation. \textsc{FlashDiffusion} thus inherits FlashAttention's tiling at the expense of one or two normalization passes per new $\beta$.

For row and column tiles $I$ and $J$, the observations $\bar{\mathcal{R}}_I$ and $\bar{\mathcal{R}}_J$, their norms $r_I,r_J$, and the Lanczos block $V_J$ are loaded from HBM into on-chip memory. The tile is evaluated as, with $\mathcal{G}_{IJ}=\bar{\mathcal{R}}_I\bar{\mathcal{R}}_J^\top$,
\begin{align}
\mathsf{K}_{IJ}&=\exp_{\odot}\left(-\beta\left( r_I\mathbf{1}^\top+\mathbf{1}r_J^\top-2\mathcal{G}_{IJ} \right)\right),
\end{align}
followed immediately by:
\begin{align}\label{eq:tiled_rbf}
Y_I \leftarrow Y_I + \mathsf{K}_{IJ}V_J .
\end{align}
The intermediate matrices $\mathcal{G}_{IJ}$ and $\mathsf{K}_{IJ}$ exist only within the working tile and are discarded after accumulation. Iterating over $J$ therefore evaluates the exact dense kernel application without ever storing the $\mathcal{O}(N^2)$ affinity matrix. This computation maps directly onto GPU streaming multiprocessors: independent output tiles are distributed across thread blocks, the local Gram product runs on the tensor cores, and the distance transform, exponential, and accumulation are fused into the same GPU kernel as its epilogue. The implementation targets NVIDIA Ampere and Blackwell architectures \citep{nvidia2020_ampereArchitecture, nvidia2024_blackwellArchitecture} using \textsc{CUTLASS}/\textsc{CuTe} primitives \citep{thakkar2023_cutlass}. Yielding \textsc{XGEMM}, an eXponential-\textsc{GEMM}\footnote{General Matrix Multiply.} primitive.

\subsection{$\beta$-function flow}\label{sec:ladder}

The matrix-free operator provides the spectrum at a fixed bandwidth $\beta$; \textsc{FlashDiffusion} additionally determines the bandwidth at which the requested spectral basis is best resolved. The algorithm treats $\log\beta$ as a one-dimensional continuation coordinate and advances through a sequence of spectral probes. Each probe constructs the normalized diffusion operator at the current $\beta$, solves for the Perron mode together with the requested nontrivial eigen-pairs using block implicitly restarted Lanczos, and returns the quantities required to determine the next step and evaluate the bandwidth-selection criterion \citep{sorensen1992_implicitlyRestartedArnoldi,baglama2003_irbl}. For centered observations, the initial bandwidth is chosen directly from the data. Since $\mathbb{E}_{ij}\lVert x_i-x_j\rVert^2=2\mathbb{E}_i\lVert x_i\rVert^2$, \textsc{FlashDiffusion} initializes:
\begin{align}\label{eq:beta0}
\beta_0=\min\left(0.1,\frac{1}{2N^{-1}\sum_{i=1}^N\lVert \bar{\mathcal{R}}_i\rVert^2}\right).
\end{align}
Here $\bar{\mathcal{R}}$ is the centered data matrix. The hardcoded cap at $0.1$ places the initial probe conservatively on the broad-kernel side when the variance-based estimate would otherwise begin at a narrow kernel. The initial scale is therefore determined in $\mathcal{O}(ND)$ work before any quadratic kernel application is required.

The potential-based continuation advances on a dyadic grid, $\beta\mapsto2\beta$. The dyadic structure is required by the effective-sample estimator below, which compares the off-diagonal kernel masses at neighboring probes $\beta$ and $2\beta$. Elementwise, the Gaussian kernel satisfies $\mathsf{K}(2\beta)=\mathsf{K}(\beta)^{\odot 2}$, so the two probes correspond exactly to adjacent scales of the same kernel family. The quantity $S_{\mathrm{off}}(2\beta)$ is accumulated during the next probe's own kernel evaluation, so evaluating the effective-sample statistic requires no kernel pass beyond those already present in the natural progression of the sweep. Neighboring probes also have slowly varying eigen-spaces, so the eigenvectors from the previous probe provide the warm start for the next using the Nyström extension \cite{nystrom1930_integralEquations}. The solver follows a gradually rotating spectral subspace rather than restarting from an uninformed initialization. A non-dyadic continuation with adaptive ratios $r_{\mathrm{fast}}=1.5$, $r_{\mathrm{slow}}=1.2$, and $r_{\mathrm{crawl}}=1.05$, selected from Lanczos cycle count and Perron-gap proximity, is retained for API compatibility but does not use the empirical potential selector described below. Bandwidth selection uses quantities already produced by normalization and the eigensolver. The first term measures the spectral heat action of the retained nontrivial modes,
\begin{align}\label{eq:vbias}
\mathcal{V}_{\mathrm{bias}}(\beta)=\operatorname{median}_{j}\left(-\log\left(\frac{\lambda_j}{\lambda_0}\right)\right).
\end{align}
As $\beta\rightarrow0$, the Gaussian affinity approaches the rank-one matrix $\mathbf{1}\mathbf{1}^{\top}$ and the nontrivial diffusion spectrum collapses, hence $\mathcal{V}_{\mathrm{bias}}\to\infty$. In the local heat-kernel regime, $\mathcal{V}_{\mathrm{bias}}(\beta)\sim\beta^{-1}$, recording the departure of the requested eigenspace from the broad-kernel rank-one limit. Finite sampling contributes a second scale through an effective sample size measured along the sweep. Let $S_{\mathrm{off}}(\beta)=\sum_{i\neq j}e^{-\beta\lVert x_i-x_j\rVert^2}$. For neighboring dyadic probes, \textsc{FlashDiffusion} defines:
\begin{align}\label{eq:neff}
n_{\mathrm{eff}}(\beta)=\frac{S_{\mathrm{off}}(\beta)^2}{N S_{\mathrm{off}}(2\beta)}.
\end{align}
For $k$ equally weighted interactions per row this reduces to $k$, while for nonuniform weights it gives the corresponding effective interaction count. In the heat-kernel regime it scales as $n_{\mathrm{eff}}\sim N\beta^{-d/2}$. The corresponding relative statistical uncertainty is:
\begin{align}\label{eq:vstat}
\mathcal{V}_{\mathrm{stat}}(\beta)=\frac{1}{\sqrt{n_{\mathrm{eff}}(\beta)\mathcal{V}_{\mathrm{bias}}(\beta)}}.
\end{align}
Because both $S_{\mathrm{off}}(\beta)$ and $S_{\mathrm{off}}(2\beta)$ are produced by consecutive probes of the dyadic sweep, the statistic requires no auxiliary kernel evaluation.
At large $\beta$, the diffusion operator approaches the identity and the leading spectral gaps eventually become comparable to the arithmetic precision of the kernel evaluation and eigensolver. Define the normalized Perron gap as $g_P(\beta)=(\lambda_0-\lambda_1)/|\lambda_0|$. \textsc{FlashDiffusion} incorporates a smooth machine-precision barrier,
\begin{align}\label{eq:vmach}
\mathcal{V}_{\mathrm{mach}}(\beta)=\left(\frac{\kappa_{\mathrm{P}}\eta}{g_P(\beta)}\right)^p,
\end{align}
with exponent $p=6$ and safety factor $\kappa_{\mathrm{P}}=10$, together with the hard admissibility condition:
\begin{align}
g_P(\beta)>\kappa_{\mathrm{P}}\eta.
\label{eq:perronwall}
\end{align}
Here $\eta$ is the numerical precision scale of the kernel data-type\footnote{With $\eta=8\times10^{-3}$ for \texttt{bf16}, $10^{-3}$ for \texttt{tf32}, $10^{-7}$ for \texttt{fp32}, and $2.22\times10^{-16}$ for \texttt{fp64}. The defaults reported here use \texttt{fp32} throughout, with $\eta=10^{-7}$. The precision cascade described below is enabled only when multiple data-types are requested.} The smooth term shapes the approach to the precision boundary from within the admissible region, while the hard condition prevents selection once the Perron and first nontrivial eigenvalues can no longer be reliably separated.
The sweep therefore follows a measured trajectory between two spectral limits. Increasing $\beta$ progressively resolves additional nontrivial modes. \textsc{FlashDiffusion} combines the heat-action, finite-sample, and machine-precision terms into a dimensionless empirical potential on $\log\beta$,
\begin{align}\label{eq:vtotal}
\mathcal{V}(\beta)=\mathcal{V}_{\mathrm{bias}}(\beta)+\mathcal{V}_{\mathrm{stat}}(\beta)+\mathcal{V}_{\mathrm{mach}}(\beta).
\end{align}
The ideal bandwidth $\beta^\star$ is the minimizer of this finite-sample resolution potential within the admissible spectral regime. The discrete sweep returns an achieved bandwidth $\beta^\times\leq\beta^\star$, determined by the probes that can be resolved safely at the available sampling and arithmetic precision. A minimum is accepted once three consecutive probes bracket a stationary point in $\log\beta$ for which $\mathcal{V}_{\mathrm{bias}}$ is falling and either $\mathcal{V}_{\mathrm{stat}}$ or $\mathcal{V}_{\mathrm{mach}}$ is rising. This direction of the flow identifies the bias-variance well associated with the resolution boundary. A quadratic fit in $\log\beta$ over the bracketing triplet estimates the vertex, which is then resolved by a final probe. If the Perron wall is reached before the bracket is completed, the last safe probe defines $\beta^\times$.

The continuation provides several geometric diagnostics without an additional kernel pass. In the local-manifold regime, the log-derivative of the off-diagonal kernel mass across a dyadic pair yields the empirical intrinsic dimension:
\begin{align}\label{eq:dhat}
\widehat d(\beta)=-2\frac{\Delta\log S_{\mathrm{off}}}{\Delta\log\beta}.
\end{align}
The corresponding Singer-type scaling exponent can then be reported diagnostically. These quantities characterize the observed flow and permit comparison with continuum asymptotics, while the selected bandwidth is determined from the measured finite-sample spectrum rather than from a prescribed value of $\widehat d$ or a hard-coded scaling law.

When multiple kernel data-types are supplied, \textsc{FlashDiffusion} executes the continuation as a precision cascade. A sweep is first performed with the earliest requested data-type, and its selected state initializes the next precision level. The final bandwidth and eigenspace are taken from the last data-type in the cascade. This separates the precision used to explore the $\beta$ trajectory from the precision required to resolve its final spectral boundary while preserving the same normalization, matrix-free kernel application, and block Lanczos machinery at every stage \citep{jax2018_jax,paszke2019_pytorch}. The default configuration uses a single data-type, \texttt{fp32}. The complete algorithm therefore consists of three nested operations: \textsc{XGEMM} supplies matrix-free Gaussian applications, block implicitly restarted Lanczos converts those applications into finite spectral probes, and the $\beta$ flow uses successive probes to approach the finite-sample resolution optimum. The output is achieved with bandwidth $\beta^\times$, and its associated nontrivial diffusion eigenvalues and the corresponding symmetric eigenvectors. From which the Markov right eigenvectors are obtained by diagonal rescaling.

%% file: sections/T6experiment.tex
\section{Experiment: $T^6$ Manifold}\label{sec:T6}

We benchmark \textsc{FlashDiffusion} on a manifold whose continuum spectrum is known in closed form. The flat six-torus $T^6 = (S^1)^6$ with unequal radii:
\begin{align}
R = (1.00,\,1.08,\,1.17,\,1.27,\,1.38,\,1.50)
\label{eq:t6radii}
\end{align}
each circle embedded as $(R_a\cos\theta_a,\,R_a\sin\theta_a)$ with independent uniform angles $\theta_a\sim\mathrm{Uniform}[0,2\pi)$, is a curvature-free product manifold in $\mathbb{R}^{12}$, zero-padded to $D=32$ and sample-centered.\footnote{Only the first twelve ambient
coordinates are non-zero; the padding fixes the ambient dimension for the
kernel without adding structure. Samples are centered by subtracting the
sample mean, which preserves the anisotropic radii and matches the centered input expected by \textsc{FlashDiffusion}. All angles are drawn from a single seed ($42$) per $N$ so that the validation diagnostic is reproducible; neither the seed nor the angles enter the solve, only the post-hoc score of Figure~\ref{fig:t6scaling}c.} The continuum Laplace--Beltrami eigenvalues are $\sum_a m_a^2/R_a^2$ with integer mode numbers $m_a$. The unequal radii break the inter-circle degeneracies, and because $R_{\max}/R_{\min}=1.50<2$ the entire first-harmonic ladder lies strictly below the first second-harmonic level: the largest first harmonic is $1/R_{\min}^2=1.00$, while the smallest second harmonic is $4/R_{\max}^2\approx1.778$, leaving a gap of $\approx0.78$ between the first-harmonic band and the rest of the spectrum. Each circle contributes a degenerate doublet $\{\cos(m\theta_a),\sin(m\theta_a)\}$ per harmonic $m$, so the first-harmonic band consists of six doublets; the leading doublet, that of the largest-radius circle ($s=5$, $R_s=1.50$), provides an exact target for the recovered modes.

We compute the leading $k=3$ nontrivial modes at $\alpha=1/2$ in \texttt{fp32} on NVIDIA RTX~PRO~6000 (Blackwell SM120) GPUs. Throughout, $\beta^\star(N)$ denotes the ideal resolution scale of \S\ref{sec:ladder}: the largest bandwidth at which the requested leading spectrum remains distinguishable from the sampling and numerical resolution floors. Section~\ref{sec:ladder} shows that $\beta^\star$ exists but is not computable from finite data; the in-loop heat-action selector returns an achievable approximation $\beta^\times(N) \le \beta^\star(N)$, subject to the Perron ceiling $\beta^\times$ that enforces the machine-precision bandwidth floor. The selector runs as it would on an unlabeled dataset, and the ground-truth angles enter only the post-hoc validation metric described below. The third computed mode ($k=3$) is retained only as a Lanczos buffer, letting the solver certify that the multiplicity-two target doublet has closed before a possible fourth mode.

The sample count varies over the power-of-two sequence $N=2^p$, up-to a dataset of $2^{22}=4{,}194{,}304$ observations. Timing variance is largest for small problems while the per-solve cost grows quadratically, so the number of repetitions decays geometrically from $64$ at the smallest $N$ to $1$ at the largest, following $n_{\mathrm{rep}}(N)=\mathrm{clip}\left(\lceil
64^{(22-\log_2 N)/12}\rceil,\,1,\,64\right)$, giving $n_{\mathrm{rep}}(2^{10})=64$, $n_{\mathrm{rep}}(2^{17})=6$, and $n_{\mathrm{rep}}(2^{22})=1$.  We report mean wall-clock time with one standard deviation over repeats.

\input{sections/T6figure}

Figure~\ref{fig:t6scaling}a shows time to solution. Below $N\approx2^{14}$ the runtime is flat at the fixed launch and solver overhead ($\approx6$\,s); above it the measured cost approaches the $\mathcal{O}(N^2)$ reference of the dense kernel application, eq.~\ref{eq:matmat}, with fitted exponent $\gamma=0.157$ over the quadratic regime. A single GPU reaches $N=2^{20}$ in $37$\,min, and four GPUs reach $N=2^{22}$ ($\approx4.2$M observations) in $3.5$\,h. A dense \texttt{fp32} affinity matrix at $N=2^{22}$ would occupy $N^2\times4\,\mathrm{bytes}
\approx7.0\times10^{13}\,\mathrm{bytes}$, or $\approx70$\,TB. At $N=2^{20}$ the four-GPU configuration completes in $781$\,s against the single-GPU $2225$\,s, a $2.85\times$ speedup or $71\%$ parallel efficiency, with multi-GPU runs sharded over the row axis.

Figure~\ref{fig:t6scaling}b reports the selector's calibrated bandwidth $\beta^\times(N)$, the achievable approximation to the ideal resolution scale $\beta^\star(N)$ of \S\ref{sec:ladder}, shown against the Singer reference slope $N^{2/(d+6)}=N^{1/6}$ for $d=6$ anchored at the first point. The reference exponent follows from the graph-to-manifold convergence of
\cite{singer2006_graphToManifoldLaplacian, hein2007_graphLaplaciansRandomNeighborhoodGraphs}, where the bandwidth must shrink with sample count. For a $d$-dimensional manifold sampled with a Gaussian kernel, the resulting bandwidth scales as $N^{-2/(d+6)}$, thus $\beta^\star$ grows as the reciprocal. Since $\beta^\times \le \beta^\star$ pointwise, the measured curve is a lower estimate of the ideal curve; the visible gap to the Singer guide can reflect either a genuine sub-Singer scaling of $\beta^\star$ or a $\beta^\star - \beta^\times$ bias whose $N$-dependence is not separately known. The measured slope over the quadratic regime is $0.157$, to be compared with $1/6\approx0.1667$; the measured curve runs somewhat shallower, $\approx20\%$ below the Singer guide at the largest $N$. A visible break in the per-$N$ error bars near $N=2^{17}$ coincides with the drop of $n_{\mathrm{rep}}$ into the single-digit regime and may also reflect a change in the selector's effective safety margin as the spectrum hardens; it does not shift the central trend of $\beta^\times(N)$.

Figure~\ref{fig:t6scaling}c reports recovery of the known leading eigenspace. Because the continuum doublet is degenerate and may rotate arbitrarily, we score the two-dimensional span rather than individual eigenvectors: the whitened cross-covariance between the leading two nontrivial modes and $\{\cos\theta_s,\sin\theta_s\}$ yields a rotation-invariant overlap in $[0,1]$, accumulated block-wise in \texttt{fp64} so that the diagnostic remains stable at multi-million $N$.\footnote{Writing $A$ for the matrix whose columns are the leading two nontrivial modes, $B$ for the columns $(\cos\theta_s,\sin\theta_s)$, and $G_{AA}$, $G_{BB}$, $G_{AB}$ for their centered covariance blocks, the score is $\sum_{ij}C_{ij}^2/2$ with $C=G_{AA}^{-1/2}G_{AB}G_{BB}^{-1/2}$, the squared RV coefficient between the two subspaces. It equals $1$ iff the subspaces coincide and $0$ iff they are orthogonal, up to floating-point round-off. The accumulation is blocked at $10^6$ rows and carried in \texttt{fp64} so that the $N\times2$ blocks are never materialized at full precision.} The subspace error: $(1-\mathrm{overlap})$ decays from $7\times10^{-2}$ at $N=2^{10}$ to $\approx10^{-4}$ at $N=2^{22}$, confirming that the in-loop calibration tracks the resolvable geometry as sampling deepens: the leading harmonic is recovered to three significant digits by the time the sample count reaches $10^6$.

%% file: sections/T6figure.tex

\pgfplotstableread[col sep=comma]{
N,ngpus,t,ts
1024,1,6.0325,4.5185
2048,1,5.8926,4.3790
4096,1,5.8238,4.4080
8192,1,5.9655,4.4799
16384,1,6.3727,4.9327
32768,1,11.5528,9.5942
65536,1,15.9784,10.9733
131072,1,61.5799,7.1353
262144,1,147.8957,2.1209
524288,1,464.2531,2.5314
1048576,1,2225.0375,3.8461
}\tOneGpu
\pgfplotstableread[col sep=comma]{
N,ngpus,t,ts
1048576,4,780.8797,16.0360
2097152,4,3198.0657,20.4661
4194304,4,12447.1016,0
}\tFourGpu
\pgfplotstableread[col sep=comma]{
N,bstar,bstars,ov,ovs
1024,0.36936,0.03992,0.92903,0.00106
2048,0.42356,0.06325,0.96819,0.00220
4096,0.49509,0.08424,0.98477,0.00159
8192,0.58220,0.10710,0.98515,0.00174
16384,0.68935,0.13137,0.99437,0.00180
32768,0.80013,0.18802,0.99428,0.00161
65536,0.91729,0.27512,0.99720,0.00145
131072,0.70902,0,0.99918,0
262144,0.78405,0,0.99954,0
524288,0.87415,0,0.99968,0
1048576,0.97949,0,0.99980,0
2097152,1.09430,0,0.99989,0
4194304,1.21354,0,0.99989,0
}\tLadder

\begin{figure}[t]
\centering
\begin{tikzpicture}
\begin{groupplot}[
  group style={group size=2 by 1, horizontal sep=2.6cm},
  width=0.48\textwidth, height=0.40\textwidth,
  xmode=log, log basis x=2,
  xlabel={$N$},
  xtick={1024,16384,262144,4194304},
  xticklabels={$2^{10}$,$2^{14}$,$2^{18}$,$2^{22}$},
  tick label style={font=\small},
  label style={font=\small},
  legend style={font=\scriptsize, draw=none, fill=none},
  every axis title/.style={font=\small, at={(0.5,1.02)}, anchor=south},
]

\nextgroupplot[
  ymode=log,
  ylabel={wall time (s)},
  title={(a) time to solution},
  legend pos=north west,
]
\addplot+[mark=*, mark size=1.7pt, error bars/.cd, y dir=both, y explicit]
  table[x=N, y=t, y error=ts]{\tOneGpu};
\addlegendentry{1 GPU}
\addplot+[mark=square*, mark size=1.7pt, error bars/.cd, y dir=both, y explicit]
  table[x=N, y=t, y error=ts]{\tFourGpu};
\addlegendentry{4 GPUs}
\addplot[dashed, gray, domain=131072:4194304, samples=2]
  {61.58*(x/131072)^2};
\addlegendentry{$\propto N^2$}

\nextgroupplot[
  ymode=log,
  ylabel={$\beta^\star$},
  title={(b) calibrated bandwidth},
  legend pos=north west,
  ymin=0.25, ymax=2.2,
]
\addplot+[mark=*, mark size=1.7pt, error bars/.cd, y dir=both, y explicit]
  table[x=N, y=bstar, y error=bstars]{\tLadder};
\addlegendentry{$\beta^\times(N)$}
\addplot[dashed, gray, domain=1024:4194304, samples=2]
  {0.36936*(x/1024)^(1/6)};
\addlegendentry{$\propto N^{2/(d+6)}=N^{1/6}$}

\end{groupplot}
\end{tikzpicture}

\vspace{2pt}

\begin{tikzpicture}
\begin{axis}[
  width=0.60\textwidth, height=0.40\textwidth,
  xmode=log, log basis x=2, ymode=log,
  xlabel={$N$},
  ylabel={$1-$ overlap},
  xtick={1024,16384,262144,4194304},
  xticklabels={$2^{10}$,$2^{14}$,$2^{18}$,$2^{22}$},
  tick label style={font=\small},
  label style={font=\small},
  title style={font=\small},
  title={(c) first-harmonic error},
]
\addplot+[mark=*, mark size=1.7pt,
  error bars/.cd, y dir=both, y explicit]
  table[x=N, y expr={1-\thisrow{ov}}, y error expr={\thisrow{ovs}}]{\tLadder};
\end{axis}
\end{tikzpicture}
\caption{FlashDiffusion on uniform samples of the flat torus $T^6$
(RTX~PRO~6000). \textbf{(a)}~Wall time versus sample count; the dashed
guide is exact $\mathcal{O}(N^2)$ scaling, and the flat region below
$N \approx 2^{14}$ is the fixed launch/solver overhead. Four-GPU
points continue the curve to $N=2^{22} \approx 4.2$M.
\textbf{(b)}~The bandwidth emitted by the $(n,\beta)$ ladder,
against the Singer exponent $2/(d+6)=1/6$ for $d=6$ (dashed, anchored at
the first point). \textbf{(c)}~Recovery of the first harmonic band:
projection error of the leading nontrivial eigenspace onto the exact
$T^6$ harmonics, decaying with sample count toward the resolution floor.
Error bars are one standard deviation over repeated runs where multiple
runs were performed.}
\label{fig:t6scaling}
\end{figure}
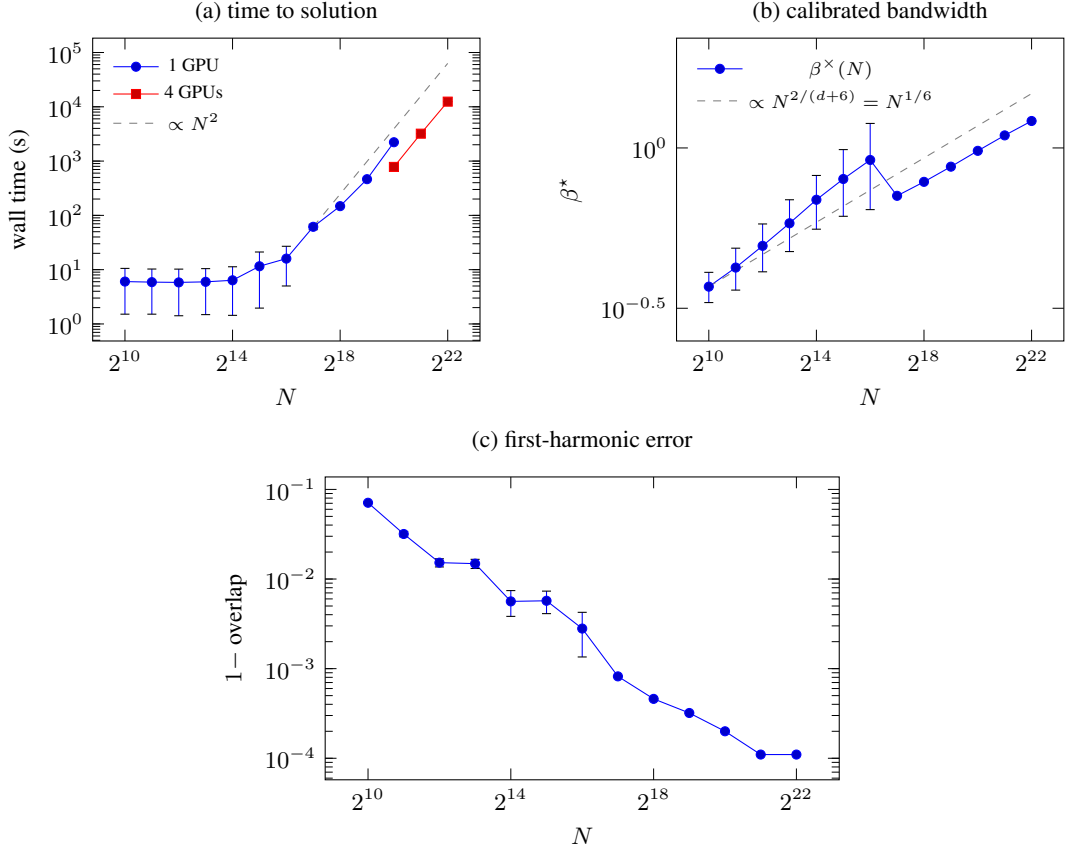

%% file: sections/AlanineDipeptide.tex
\section{Experiment: Alanine Dipeptide}\label{sec:alanine}

We evaluate \textsc{FlashDiffusion} on the \textsc{MDShare} alanine-dipeptide benchmark \citep{nueske2017_markovStateModels,wehmeyer2018_timeLaggedAutoencoders}, consisting of three independent $250$ ns molecular-dynamics trajectories and $N=750{,}000$ configurations. Each configuration contains the $10$ heavy atoms and is mapped to an E(3)-invariant representation using distances from every heavy atom to the first $8$ heavy atoms in the dataset ordering as anchors \citep{candanedo2025_e3InvariantBiomolecules}, giving $D=80$ features. \textsc{FlashDiffusion} is run with $\alpha=1/2$ in \texttt{fp32} and computes $256$ nontrivial diffusion modes on eight NVIDIA RTX PRO 6000 GPUs. The eigensolve required $7315$ s ($2.03$ h), and explicit storage of the corresponding dense \texttt{fp32} kernel would require $750{,}000^2\times4\approx2.25$ TB, which is once again avoided by the tiled matrix-free operator. This experiment uses the full $N$ throughout and exercises the $\beta$ continuation directly rather than the sample-size ladder.

\input{figures/ramachandran_updated}

\input{figures/spectralres_updated}

The $\beta$ flow required eight dyadic probes and selected $\beta^\star=276.4$ from the empirical potential. The sweep continued beyond the selected point to characterize the precision boundary: $\beta=552.8$ remained admissible, while $\beta=1105.7$ crossed the \texttt{fp32} Perron wall with $g_P=3.51\times10^{-9}$. At $\beta^\star$, the effective interaction count is $n_{\mathrm{eff}}\approx4.31\times10^4$, approximately $5.7\%$ of the sample count, and the measured local dimension is $\widehat d\approx4.85$. Here $\widehat d$ is a kernel-scale diagnostic of the resolved geometry in the E(3)-invariant representation and should not be identified with the two-dimensional $(\phi,\psi)$ reaction coordinate alone. The selected $\beta^\star$ therefore lies comfortably inside the resolved spectral regime rather than at the \texttt{fp32} precision boundary.
Following the diffusion-map autoencoder construction of \citet{candanedo2024_diffusionMapAutoencoder}, we fit a single linear decoder from the first $m$ diffusion coordinates to the circular targets $(\cos\phi,\sin\phi,\cos\psi,\sin\psi)$. The decoder is fit by least squares with ridge regularization $10^{-8}\operatorname{tr}(\mathcal{G}^{(m)})/m$, where $\mathcal{G}^{(m)}$ is the centered decoder Gram matrix. Two trajectories supply the decoder-fit population, from which $5,000$ frames are drawn uniformly without replacement as an additional random holdout, while the third $250$ ns trajectory is excluded entirely from decoder fitting. The diffusion eigensolve itself remains unsupervised over all configurations; only the decoder uses the split. We evaluate $m\in\{2,4,8,16,32,64,128,256\}$ using the mean absolute wrapped angular error $\big|\arg\exp(i(\widehat\theta-\theta))\big|$.
Figure~\ref{fig:alanine} shows distinct resolution profiles for the two backbone coordinates. The held-out $\phi$ error falls from $26.69^\circ$ at $m=2$ to $7.71^\circ$ at $m=8$, while $\psi$ remains near $15^\circ$ through $m=8$ and begins its principal decrease at $m_\psi=16$. At $m=256$, the completely held-out trajectory reaches $4.86^\circ$ MAE for $\phi$ and $3.95^\circ$ for $\psi$, compared with $4.90^\circ$ and $3.96^\circ$ on the random holdout. The differences are below $0.05^\circ$, so the fully held-out trajectory shows no observable degradation relative to frames drawn from the decoder-training trajectories at the reported precision. Reconstruction also continues beyond the ambient feature dimension $D=80$: from $m=128$ to $m=256$, the held-out errors decrease from $5.26^\circ$ to $4.86^\circ$ for $\phi$ and from $4.42^\circ$ to $3.95^\circ$ for $\psi$. Figure~\ref{fig:alanine} additionally compares the held-out Ramachandran surface with decoded ensembles at $m=8$ and $m=256$; the latter recovers the principal metastable regions and their geometry, while a strided held-out trajectory segment has mean joint angular error $7.1^\circ$. The spectrum confirms that these additional modes remain part of a substantial high-rank operator. All $256$ computed eigenvalues are finite, positive, and correctly ordered, with $\lambda_{\max}=0.9985$ and $\lambda_{256}=0.2127$. The final computed eigenvalue therefore remains approximately $21\%$ of the leading value and lies more than six orders of magnitude above the \texttt{fp32} working precision scale used by the sweep. Together with the continued reconstruction improvement for $m>D$, this shows that the useful diffusion representation extends well beyond a low-rank approximation of the $80$-dimensional input features and provides a molecular example of the high-rank Gaussian regime targeted by \textsc{FlashDiffusion}.

%% file: figures/ramachandran_updated.tex
%
%

\begin{figure}[t]
\centering
\begin{tikzpicture}
\begin{groupplot}[
  group style={
    group size=3 by 1,
    horizontal sep=0.45cm,
    ylabels at=edge left,
    yticklabels at=edge left
  },
  scale only axis,
  width=0.247\textwidth,
  height=0.279\textwidth,
  xmin=-180,
  xmax=180,
  ymin=-180,
  ymax=180,
  xtick={-180,0,180},
  ytick={-180,0,180},
  xticklabels={$-180^{\circ}$,$0^{\circ}$,$180^{\circ}$},
  yticklabels={$-180^{\circ}$,$0^{\circ}$,$180^{\circ}$},
  xlabel={$\psi$},
  ylabel={$\phi$},
  axis on top,
  axis lines=box,
  tick align=outside,
  tick label style={font=\scriptsize},
  label style={font=\small},
  ylabel style={xshift=0.18cm},
  title style={font=\small},
  enlargelimits=false,
  clip=false
]

\nextgroupplot[title={(a) Reference}]
\addplot graphics[
  xmin=-180,xmax=180,
  ymin=-180,ymax=180
] {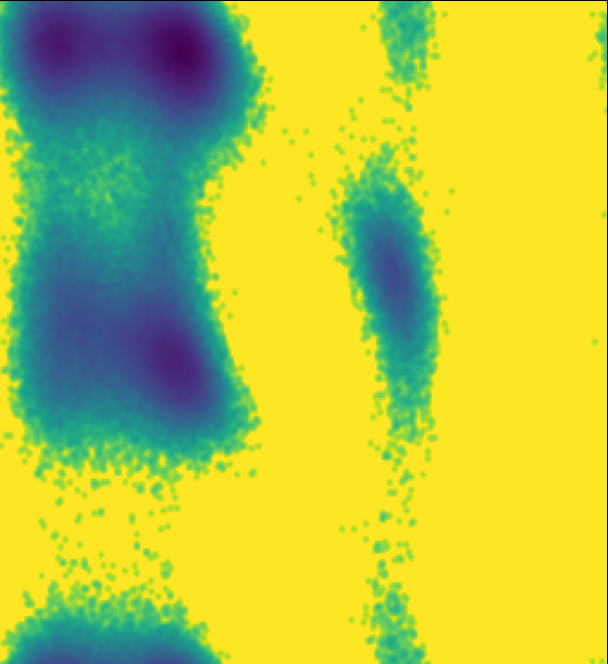};

\nextgroupplot[title={(b) 8 DMAP modes}]
\addplot graphics[
  xmin=-180,xmax=180,
  ymin=-180,ymax=180
] {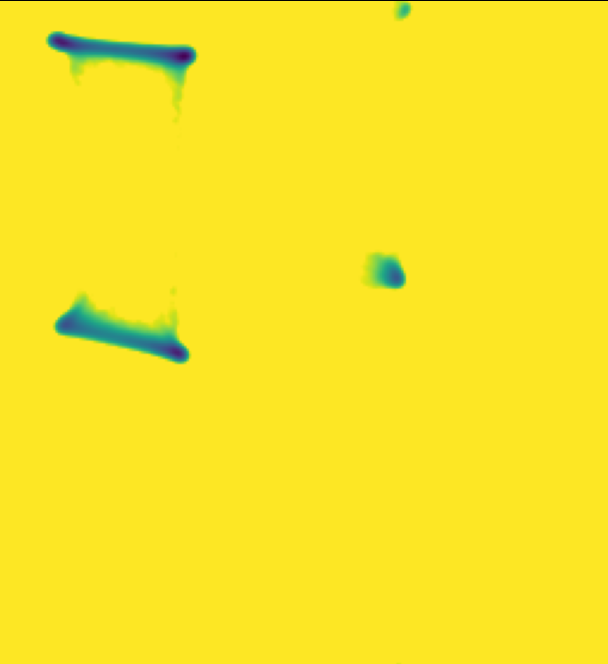};

\nextgroupplot[title={(c) 256 DMAP modes}]
\addplot graphics[
  xmin=-180,xmax=180,
  ymin=-180,ymax=180
] {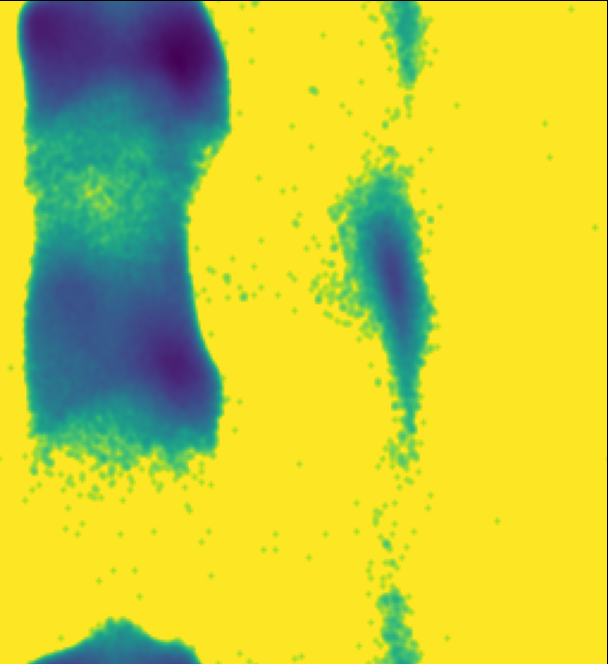};

\end{groupplot}

\begin{axis}[
  at={(group c3r1.east)},
  anchor=west,
  xshift=0.18cm,
  scale only axis,
  width=0.015\textwidth,
  height=0.279\textwidth,
  xmin=0,
  xmax=1,
  ymin=0,
  ymax=8,
  xtick=\empty,
  ytick={0,1,2,3,4,5,6,7,8},
  axis x line=none,
  axis y line*=right,
  tick align=outside,
  tick label style={font=\scriptsize},
  ylabel={$F/k_B T$},
  label style={font=\small},
  ylabel style={font=\small, rotate=180, yshift=-0.10cm},
  enlargelimits=false,
  clip=false,
  axis on top
]
\addplot graphics[
  xmin=0,xmax=1,
  ymin=0,ymax=8
] {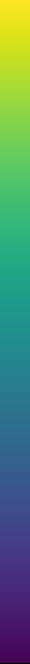};
\end{axis}

\end{tikzpicture}
\caption{Ramachandran free-energy distributions for alanine dipeptide. \textbf{(a)} Reference distribution, \textbf{(b)} reconstruction using 8 diffusion-map (DMAP) modes, and \textbf{(c)} reconstruction using 256 DMAP modes. The horizontal coordinate is the dihedral angle $\psi$ and the vertical coordinate is $\phi$; both span $[-180^{\circ},180^{\circ}]$, with $(\phi,\psi)=(0,0)$ at the center of each panel. The shared color bar reports the free energy $F/k_B T$ from 0 to 8.}
\label{fig:alanine-ramachandran}
\end{figure}

%% file: figures/spectralres_updated.tex
%

\begin{figure}[t]
\centering

\definecolor{spectralBlue}{RGB}{31,119,180}
\definecolor{spectralOrange}{RGB}{255,127,14}
\definecolor{spectralGreen}{RGB}{44,160,44}
\definecolor{spectralRed}{RGB}{214,39,40}

\pgfplotstableread[col sep=comma]{
m,phi_mae_traj,psi_mae_traj,joint_traj,phi_mae_rand,psi_mae_rand,joint_rand,cond_Gm
2,26.6896,15.5255,33.4886,27.0371,15.4093,33.6612,1.23099
4,12.0095,15.3419,21.4756,12.7291,15.2708,21.9019,1.27258
8,7.70748,14.9427,18.3073,7.92088,14.9675,18.443,3604.03
16,6.7314,9.64142,13.0685,6.93115,9.69588,13.2252,10380.6
32,5.93265,6.71444,9.98026,6.04413,6.65027,9.97478,18140.9
64,5.65941,5.13687,8.48853,5.75153,5.10989,8.5287,84775.1
128,5.26102,4.42413,7.60024,5.328,4.45284,7.67385,132625
256,4.85845,3.94828,6.91114,4.89654,3.95931,6.94705,185252
}\spectralResolutionData

\begin{tikzpicture}
\begin{groupplot}[
  group style={
    group size=2 by 1,
    horizontal sep=1.55cm,
  },
  height=0.40\textwidth,
  tick label style={font=\scriptsize},
  label style={font=\small},
  title style={font=\small},
]

\nextgroupplot[
  width=0.465\textwidth,
  xmode=log,
  log basis x=2,
  xmin=1.7,
  xmax=300,
  ymin=3,
  ymax=28.5,
  xtick={2,4,8,16,32,64,128,256},
  xticklabels={$2^1$,$2^2$,$2^3$,$2^4$,$2^5$,$2^6$,$2^7$,$2^8$},
  xlabel={Number of diffusion modes $m$},
  ylabel={Held-out angular MAE (deg)},
  title={(a) Circular decoding at $\beta^\star=276.4$},
  legend style={
    at={(0.985,0.985)},
    anchor=north east,
    font=\tiny,
    draw=iclrTextColor!35,
    fill=iclrPageColor,
    fill opacity=0.90,
    text opacity=1,
    cells={anchor=west},
    legend columns=1,
    row sep=-1pt,
    inner xsep=2pt,
    inner ysep=2pt,
  },
]

\addplot+[
  spectralBlue,
  solid,
  mark=*,
  mark size=1.7pt,
  thick,
] table[x=m,y=phi_mae_traj]{\spectralResolutionData};
\addlegendentry{$\phi$ (held-out traj.)}

\addplot+[
  spectralOrange,
  solid,
  mark=*,
  mark size=1.7pt,
  thick,
] table[x=m,y=psi_mae_traj]{\spectralResolutionData};
\addlegendentry{$\psi$ (held-out traj.)}

\addplot+[
  spectralGreen,
  dashed,
  mark=square*,
  mark size=1.6pt,
  thick,
] table[x=m,y=phi_mae_rand]{\spectralResolutionData};
\addlegendentry{$\phi$ (random holdout)}

\addplot+[
  spectralRed,
  dashed,
  mark=square*,
  mark size=1.6pt,
  thick,
] table[x=m,y=psi_mae_rand]{\spectralResolutionData};
\addlegendentry{$\psi$ (random holdout)}

\draw[spectralOrange, densely dotted, thick]
  (axis cs:16,3) -- (axis cs:16,28.5)
  node[pos=0.50, anchor=south, rotate=90, font=\scriptsize, xshift=-1pt]
  {$m_\psi=16$};

\draw[iclrTextColor!60, densely dotted, thick]
  (axis cs:80,3) -- (axis cs:80,28.5)
  node[pos=0.08, anchor=west, rotate=90, font=\scriptsize]
  {$D=80$};

\nextgroupplot[
  width=0.405\textwidth,
  axis on top,
  xmin=-180,
  xmax=180,
  ymin=-180,
  ymax=180,
  xtick={-150,-100,-50,0,50,100,150},
  ytick={-150,-100,-50,0,50,100,150},
  xlabel={$\phi$ (deg)},
  ylabel={$\psi$ (deg)},
  ylabel style={at={(axis description cs:-0.105,0.5)}, anchor=south},
  title={(b) Held-out Ramachandran surface},
  enlargelimits=false,
]

\addplot graphics[
  xmin=-180,
  xmax=180,
  ymin=-180,
  ymax=180,
] {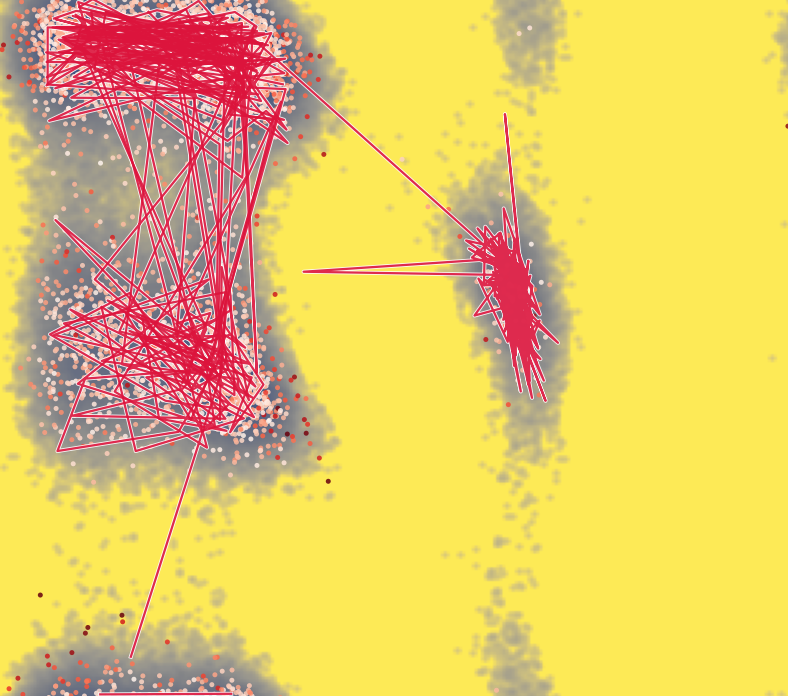};

\end{groupplot}


\begin{axis}[
  hide axis,
  scale only axis,
  width=0pt,
  height=0pt,
  point meta min=0,
  point meta max=30,
  colormap={jointErrorMap}{
    rgb255(0cm)=(255,245,240);
    rgb255(1cm)=(252,174,145);
    rgb255(2cm)=(251,106,74);
    rgb255(3cm)=(203,24,29);
    rgb255(4cm)=(103,0,13)
  },
  colorbar,
  colorbar style={
    at={(group c2r1.east)},
    anchor=west,
    xshift=0.72cm,
    width=0.18cm,
    height=0.355\textwidth,
    ytick={0,5,10,15,20,25,30},
    ylabel={joint error (deg)},
    ylabel style={font=\scriptsize},
    tick label style={font=\scriptsize},
  },
]
\addplot[draw=none] coordinates {(0,0) (1,30)};
\end{axis}

\begin{axis}[
  hide axis,
  scale only axis,
  width=0pt,
  height=0pt,
  point meta min=0,
  point meta max=8,
  colormap={freeEnergyMap}{
    rgb255(0cm)=(110,110,110);
    rgb255(1cm)=(83,92,112);
    rgb255(2cm)=(62,83,130);
    rgb255(3cm)=(49,104,142);
    rgb255(4cm)=(34,144,140);
    rgb255(5cm)=(53,183,121);
    rgb255(6cm)=(144,214,67);
    rgb255(7cm)=(220,227,24);
    rgb255(8cm)=(253,231,37)
  },
  colorbar,
  colorbar style={
    at={(group c2r1.east)},
    anchor=west,
    xshift=2.20cm,
    width=0.18cm,
    height=0.355\textwidth,
    ytick={0,1,2,3,4,5,6,7,8},
    ylabel={$F/k_B T$},
    ylabel style={font=\scriptsize},
    tick label style={font=\scriptsize},
  },
]
\addplot[draw=none] coordinates {(0,0) (1,8)};
\end{axis}

\end{tikzpicture}

\caption{\textbf{Spectral resolution of alanine-dipeptide coordinates.}
\textbf{(a)} Mean absolute wrapped error for the backbone dihedrals $\phi$ and
$\psi$ as the number of retained diffusion modes $m$ increases. Solid curves
evaluate the completely held-out trajectory, while dashed curves evaluate
$5{,}000$ randomly held-out frames from the decoder-training trajectories. The
vertical markers indicate the ambient feature dimension $D=80$ and the detected
onset $m_\psi=16$ of the principal $\psi$-resolution regime.
\textbf{(b)} Held-out Ramachandran free-energy surface at
$\beta^\star=276.4$, with random-holdout frames colored by joint angular
reconstruction error and a strided segment of the decoded held-out trajectory
overlaid as a curve.}
\label{fig:alanine}
\end{figure}
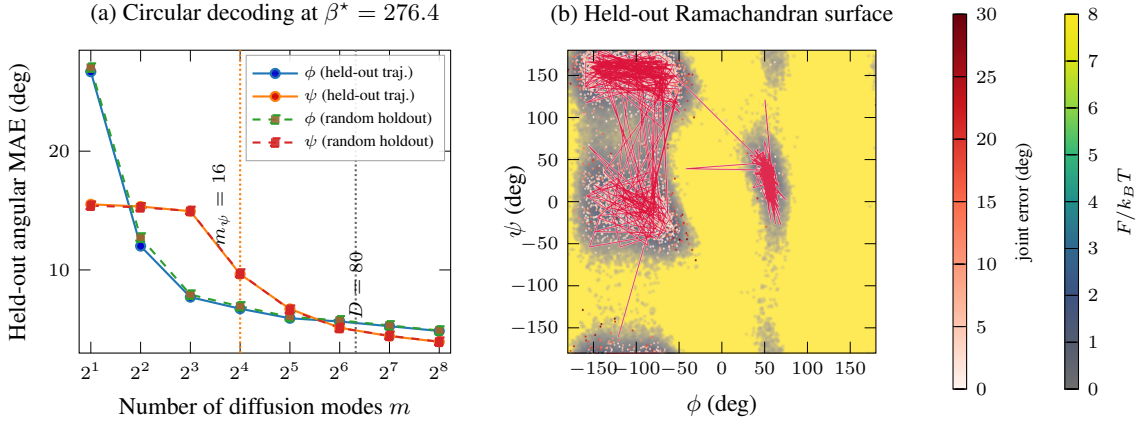

%% file: sections/Conclusion.tex
\section{Conclusion}\label{conclusion}

\textsc{FlashDiffusion} makes the pragmatic choice to retain the dense $\mathcal{O}(N^2)$ Gaussian interaction rather than replace it by a sparse or low-rank surrogate. GPU tiling removes the need to materialize the kernel, alternatives that achieve this exist in the literature including \textsc{KeOps} \cite{feydy2020_fastGeometricLearning} and we run a comparison with our primitive \textsc{XGEMM} in \S\ref{sec:keops}. However, unlike previous work we consider a synergistic connection between $\beta$-flow and tiling requirements, via the $(n,\beta)$ continuation introduced in \S\ref{sec:ladder}, to reduce the number of expensive full-data spectral solves. 

This targets the regime in which the Gaussian operator itself is high rank. Both fixed-$k$NN graphs and low-rank approximations remain effective when only a small spectral subspace is required. For data whose useful spectral rank is unknown in advance, \textsc{FlashDiffusion} instead treats rank and spectral resolution as quantities to be measured from the finite novel sample.

The $\beta$ flow also admits a scale interpretation related to renormalization ideas \citep{zinnjustin2019_randomWalksRandomMatrices}. For a Gaussian heat kernel, the diffusion time satisfies $t\propto\beta^{-1}$, hence small $\beta$ corresponds to a long-time, coarse description and large $\beta$ to progressively shorter-time, finer structure. \textsc{FlashDiffusion} follows this scale dependence in the coarse-to-fine direction as sampling increases. The ideal resolution scale $\beta^\star(N)$ characterizes the finite-sample optimum, while the algorithm returns an achievable $\beta^\times$ constrained by sampling and arithmetic precision. The accompanying diagnostics $n_{\mathrm{eff}}$ and $\widehat d$ describe how the effective neighborhood size and local spectral dimension evolve along the same flow.

The principal remaining computational limitation is the feature dimension. Each Gaussian application costs $\mathcal{O}(N^2D)$, and therefore becomes cubic when $D\sim N$. The experiments considered here remain well inside the quadratic regime, with $D/N\approx8\times10^{-6}$ on $T^6$ and $D/N\approx10^{-4}$ for alanine dipeptide. Since the kernel depends on the representation through pairwise distances, feature compression provides a natural front end: PCA, Johnson--Lindenstrauss projections \citep{johnson1984_extensions}, or learned encoders can reduce $D$ provided the induced distance distortion is controlled. This suggests a useful division of labor: the front end determines the geometry presented to the kernel, while diffusion maps perform harmonic analysis on that geometry. Jointly calibrating representation distortion and spectral resolution is a natural direction for future work.


Future work includes theoretical guarantees relating the empirical potential minimum to $\beta^\star$, joint calibration of feature-space compression and diffusion resolution, extensions to other radial kernel families, and applications where high-rank spectral structure is expected to be essential, e.g. in molecular dynamics, \cite{gowers2016_mdanalysis}. 


%% file: sections/ANN.tex
\section{Spectral Consequences of kNN/ANN Sparsification}\label{sec:knn}

A common strategy for scaling diffusion maps is $k$-nearest-neighbor ($k$NN) sparsification: retain a fixed number of neighbors per sample, store the resulting weighted graph, and apply a sparse eigensolver. This strategy is representative of many large-scale spectral-embedding pipelines that combine approximate neighborhood graphs with iterative solvers, such as \textsc{megaman} \citep{mcqueen2016_megaman}. The comparison becomes more consequential when the objective is to recover a substantial spectral basis instead of only a low-dimensional visualization. The applications considered here use on the order of $M\sim10^2$ modes. As discussed in the main text, at the bandwidths relevant to this regime the Gaussian kernel generally has substantial effective rank, which limits the effectiveness of very low-rank approximations. The matrix-free formulation therefore preserves the dense Gaussian operator while avoiding both explicit matrix construction and a fixed sparse neighborhood representation.

The first few eigenfunctions typically vary on spatial scales comparable to the manifold diameter and can therefore be comparatively insensitive to moderate changes in local graph construction. Higher eigenfunctions vary on progressively shorter scales. Weyl counting gives a characteristic wavelength of order $(V/M)^{1/d}$ for the $M$-th mode on a $d$-dimensional manifold of volume $V$. Consequently, a neighborhood graph that gives a visually faithful two- or three-dimensional embedding may preserve substantially less information than is required for a larger spectral basis. We distinguish two questions below. First, how many neighbors are required for a truncated graph to reproduce the numerically relevant support of the dense Gaussian operator at a prescribed tolerance? Second, what structural changes does hard neighborhood truncation introduce when the retained neighborhood is much smaller than this support scale? The first question leads to a small-ball scaling law, while the second is addressed through retained row mass, PSD violation, and finite-sample spectral considerations.

\subsection{Arithmetic and neighborhood truncation}\label{sec:knn:floors}

Consider the dense Gaussian kernel $\mathsf{K}_{ij}(\beta)=\exp\left(-\beta\lVert x_i-x_j\rVert^2\right)$. Although every entry is strictly positive in exact arithmetic, numerical computation introduces a practical relative scale below which individual contributions become increasingly insignificant compared with $O(1)$ accumulated quantities. We therefore introduce a working tolerance $\tau$ and regard interactions satisfying $\mathsf{K}_{ij}(\beta)<\tau$ as negligible at the requested arithmetic precision. We use $\tau\approx10^{-7}$ as an \texttt{fp32}-scale working tolerance, comparable to the relative machine-epsilon scale, and $\tau\approx10^{-15}$ as an \texttt{fp64}-scale tolerance. The resulting estimates depend only logarithmically on $\tau$. The corresponding support radius is:
\begin{align}
D_\tau=\sqrt{\frac{\log(1/\tau)}{\beta}}.
\label{eq:Dtau}
\end{align}
Interactions outside a ball of radius $D_\tau$ therefore have weight below the prescribed working tolerance.
A $k$NN construction imposes a different truncation at the $k$-th-neighbor distance $R_k(x_i)$. This radius is row dependent and varies with the local sampling density $q(x_i)$. For the quantitative estimates below we first consider uniform sampling, $q(x)\equiv q=1/V$, and let $\omega_d$ denote the volume of the unit $d$-ball. Locally, $k$ neighbors occupy a volume of order $k/(Nq)$.
For fixed $\beta$, let $\widetilde{\mathsf{K}}_{\beta,k}=\mathsf{K}_{\beta}\odot A_k$ denote the symmetrically masked kernel, with $A_k$ the neighborhood mask. Reproducing all dense-kernel interactions above tolerance $\tau$ requires approximately $R_k(x_i)\gtrsim D_\tau$ for every row. Under nonuniform sampling, this condition becomes spatially heterogeneous, so a single fixed $k$ generally corresponds to different effective support tolerances across the manifold.
This distinction is also relevant to bandwidth selection. The neighborhood size required to reproduce the dense operator at tolerance $\tau$ depends on $\beta$, while the preferred bandwidth $\beta^\star$ is determined by the spectral calculation. 

\textbf{A fixed neighborhood graph therefore selects a truncation scale before the bandwidth of interest is known.} 

FlashDiffusion evaluates the Gaussian operator directly throughout the $\beta$ continuation, so the same pairwise interaction model is used during bandwidth selection and eigenspace computation.

\subsection{Two scales: spectral resolution and kernel support}\label{sec:knn:rank}

Two distinct length scales govern the comparison. The first is the Gaussian smoothing scale $\ell_\beta\sim\beta^{-1/2}$. By Weyl counting, the number of spatial modes supported at this scale behaves as $M(\beta)\sim V\beta^{d/2}$ up to geometric constants. Resolving $M$ modes therefore corresponds to the scaling:
\begin{align}
\beta\gtrsim\left(\frac{M}{V}\right)^{2/d}.
\label{eq:betaM}
\end{align}
The second is the support scale $D_\tau$ in eq.~\ref{eq:Dtau}. Matching the dense Gaussian kernel above the working tolerance $\tau$ requires approximately:
\begin{align}
k^\ast(\beta,\tau)=\frac{N\omega_d}{V}\left(\frac{\log(1/\tau)}{\beta}\right)^{d/2}.
\label{eq:kstar}
\end{align}
The quantity $k^\ast$ is therefore a support-fidelity scale: it is the expected number of kernel interactions per row whose weights exceed the chosen tolerance.
For the Swiss roll at $N=10^6$, $\beta^\star=15$, and $d=2$, this estimate gives $k^\ast\approx8\times10^4$ for $\tau=10^{-7}$ and $k^\ast\approx1.7\times10^5$ for $\tau=10^{-15}$. The smaller \texttt{fp64}-scale tolerance includes a wider range of weak interactions and therefore increases the support count.
These values are much larger than neighborhood sizes such as $k\in{32,64,128}$ commonly used to construct sparse embedding graphs. The comparison establishes that a small fixed-$k$ graph is a strong truncation of the finite-sample Gaussian operator at $\beta^\star$. Preservation of a particular leading eigenspace can still occur under stronger truncation because spectral accuracy also depends on the structure of the omitted interactions and on the relevant eigengaps.
Along a calibrated bandwidth path, two reference exponents are useful. Balancing bias against pointwise fluctuation gives Singer's scaling $\beta^\star\sim N^{2/(d+6)}$ \citep{singer2006_graphToManifoldLaplacian}, while balancing against the spectral fluctuation $(N\varepsilon^{d/2})^{-1/2}$ gives $\beta^\star\sim N^{2/(d+4)}$, the corresponding rate for a spectral selection criterion \citep{calder2022_spectralConvergenceGraphLaplacians}. On the Swiss roll, the measured bandwidth selection follows $\beta^\star\sim N^{1/3}$ for $d=2$ over the tested range. Substitution into eq.~\ref{eq:kstar} gives $k^\ast\sim N^{4/(d+4)}$, or $k^\ast\sim N^{2/3}$ for $d=2$, consistent with the measured support counts.
Combining eq.~\ref{eq:betaM} with eq.~\ref{eq:kstar} gives a useful scaling relation. At the bandwidth associated with $M$ resolved modes, preserving the dense-kernel support above tolerance $\tau$ requires:
\begin{align}
k\gtrsim\frac{N\log(1/\tau)^{d/2}}{M}.
\label{eq:krequired}
\end{align}
Order-one geometric constants are omitted in eq.~\ref{eq:krequired}. For $d=2$ and $N=10^6$, this support-matching scale is approximately $10^5$ neighbors for $M=100$ and $10^4$ neighbors for $M=1000$ at the working tolerances considered here. A fixed graph with tens or hundreds of neighbors therefore corresponds to a substantially stronger truncation than retaining the numerically relevant support of the dense kernel. The same relation can be viewed at fixed $k$. For $d=2$, the support count scales as $\beta^{-1}$. Using the measured Swiss-roll reference point, $k^\ast\approx8\times10^4$ at $\beta^\star=15$, a graph with $k=64$ reaches a comparable support radius around:
\begin{align}
\beta_{\mathrm{support}}\approx15\left(\frac{8\times10^4}{64}\right)\approx2\times10^4.
\end{align}
This value is roughly three orders of magnitude larger than the selected $\beta^\star$. A $k=64$ graph therefore corresponds to a much shorter geometric interaction scale than the dense Gaussian operator selected by the spectral criterion.

\subsection{Numerically relevant support and retained row mass}\label{sec:knn:support}

The support calculation above follows directly from the small-distance distribution of points on a smooth manifold. Define $\rho_\tau(\beta)=\Pr[\mathsf{K}_{ij}(\beta)\ge\tau]$. Equivalently, $\rho_\tau(\beta)=\Pr[d^2\le\log(1/\tau)/\beta]$. If $F$ denotes the CDF of pairwise squared distances, then $\rho_\tau(\beta)=F(\log(1/\tau)/\beta)$.
While the bulk of $F$ depends on global manifold geometry, its small-distance behavior is local. For a smooth $d$-manifold of volume $V$ under approximately uniform sampling, the interior small-ball law gives $\Pr[d\le R]=(\omega_dR^d/V)(1+\mathcal{O}(R^2))$. Hence
\begin{align}
\rho_\tau(\beta)=\frac{\omega_d}{V}\left(\frac{\log(1/\tau)}{\beta}\right)^{d/2}\left(1+\mathcal{O}(D_\tau^2)\right).
\label{eq:densitylaw}
\end{align}
The expected support count per row is $n_\tau(\beta)=N\rho_\tau(\beta)$, and eq.~\ref{eq:kstar} follows directly from this relation.
Along the spectral scaling $\beta^\star\sim N^{2/(d+4)}$, the relative support density decreases with sample size while the absolute number of relevant interactions grows as $n_\tau(\beta^\star)\sim N^{4/(d+4)}$. The Gaussian operator therefore becomes increasingly localized in relative density while the number of interactions per row associated with a fixed relative tolerance continues to increase.
We verified the $d=2$ scaling on a unit-variance Swiss roll using $\tau=10^{-7}$. The prediction is $\rho_\tau(\beta)\propto\beta^{-1}$. Across $N=10^4$--$10^8$, the measured product $\rho_\tau(\beta^\star)\beta^\star$ remains approximately constant at $1.19$, within $1\%$ of $\omega_2\log(1/\tau)/V$. Over the same range, the selected bandwidth follows $\beta^\star\sim N^{1/3}$ and the measured support count increases from approximately $3.7\times10^3$ to $1.7\times10^6$ interactions per row, consistent with $N^{2/3}$ scaling, while the relative support density decreases from $37\%$ to $1.7\%$.
Curvature corrections enter at relative order $\mathcal{O}(D_\tau^2)$ in the interior, while boundary effects occur within a strip of width $\mathcal{O}(D_\tau)$. The observed agreement over the tested range indicates that the leading small-ball scaling captures the dominant dependence relevant to these support estimates.
Support counts depend on the chosen tolerance $\tau$. A complementary quantity with no dependence on $\tau$ is the fraction of Gaussian row mass retained by the $k$NN mask. Because the Gaussian weights are nonnegative, define
\begin{align}
f_k=\frac{\sum_{j\in\mathcal{N}_k(i)}\exp(-\beta d_{ij}^2)}{\sum_j\exp(-\beta d_{ij}^2)}.
\end{align}
Under the local uniform-density approximation,
\begin{align}\label{eq:rowmass}
f_k=\frac{\gamma(d/2,\beta R_k^2)}{\Gamma(d/2)}.
\end{align}
Since $R_k\sim(k/(Nq\omega_d))^{1/d}$, the argument scales as $\beta R_k^2\propto\beta(k/N)^{2/d}$. Along $\beta^\star\sim N^{2/(d+4)}$, fixed $k$ therefore gives $f_k\sim kN^{-4/(d+4)}$ in the small-argument regime.
For $d=2$, eq.~\ref{eq:rowmass} reduces locally to $f_k=1-e^{-c_2k}$, where $c_2$ contains the sampling-density and bandwidth factors. This relation is geometric and independent of the working tolerance $\tau$.
Table~\ref{tab:rowmass} verifies the prediction on the Swiss roll. A fitted $c_2\approx6.9\times10^{-4}$ reproduces the measured retained fractions to within approximately $0.3\%$ over the tested range.
\begin{table}[t]
\centering
\vspace{0.5em}
\small
\begin{tabular}{@{}rcc@{}}
\toprule
$k$ & retained $f_k$ & $1-e^{-c_2k}$ \\
\midrule
32 & 2.2\% & 2.18\% \\
64 & 4.4\% & 4.31\% \\
128 & 8.6\% & 8.43\% \\
256 & 16.4\% & 16.2\% \\
1024 & 50.4\% & 50.7\% \\
\bottomrule
\end{tabular}
\caption{Gaussian kernel row mass retained under fixed-$k$ truncation on a unit-variance Swiss roll with $N=2\times10^5$ and $\beta^\star=8.75$, averaged over 400 rows. The local $d=2$ prediction $1-e^{-c_2k}$ uses $c_2\approx6.9\times10^{-4}$.}
\label{tab:rowmass}
\end{table}
At this problem size, $k=64$ retains about $4.4\%$ of the Gaussian row mass at $\beta^\star$. Spectral error depends additionally on the organization of the omitted interactions and on the relevant spectral gaps, so the retained-mass fraction and eigenspace error need not coincide numerically. The small retained fraction nevertheless shows that fixed-$k$ masking at this scale produces a substantial perturbation of the dense operator.
This distinction is important for bandwidth selection. FlashDiffusion seeks the largest bandwidth for which a requested spectral subspace remains distinguishable from sampling and numerical fluctuations. Replacing $\mathsf{K}_\beta$ by $\widetilde{\mathsf{K}}_{\beta,k}$ changes the operator, its degree normalization, and its spectral gaps. The sparsified family therefore has its own finite-sample resolution boundary, which can occur at a different $\beta$ from the corresponding boundary of the dense Gaussian family.

\subsection{$k$NN truncation and PSD}\label{sec:knn:psd}

The dense Gaussian Gram matrix satisfies $\mathsf{K}_\beta\succeq0$. For a Hadamard-truncated kernel $\widetilde{\mathsf{K}}_{\beta,k}=\mathsf{K}_\beta\odot A_k$, the Schur product theorem guarantees positive semidefiniteness when $A_k\succeq0$. A symmetrized $k$NN adjacency mask generally does not satisfy this condition. Once indefiniteness is introduced, positive diagonal congruence preserves its inertia. For any positive diagonal matrix $D$, Sylvester's law of inertia gives
\begin{align}
\mathrm{inertia}\left(D(\mathsf{K}_\beta\odot A_k)D\right)=\mathrm{inertia}\left(\mathsf{K}_\beta\odot A_k\right).
\end{align}
The symmetric positive-diagonal rescalings appearing in diffusion-map normalization therefore preserve any negative eigenvalues introduced by the mask.
Table~\ref{tab:psd} illustrates this effect on the Swiss roll at a problem size for which the complete spectrum can be computed directly.
\begin{table}[t]
\centering
\vspace{0.5em}
\small
\begin{tabular}{@{}rccc@{}}
\toprule
$k$ & $\lambda_{\min}(\mathsf{K}\odot A_k)$ & neg. spectral mass & $\lambda_{\min}(A_k)$ \\
\midrule
16 & $-4.10$ & 26.9\% & $-4.79$ \\
32 & $-5.84$ & 27.8\% & $-7.85$ \\
64 & $-7.71$ & 25.8\% & $-13.44$ \\
128 & $-7.58$ & 19.0\% & $-33.07$ \\
256 & $-2.73$ & 9.0\% & $-61.03$ \\
\bottomrule
\end{tabular}
\caption{Spectrum of a symmetrically $k$NN-truncated Gaussian kernel on the Swiss roll with $N=3000$, $\beta^\star=2.158$, and neighborhood masks symmetrized by union. Negative spectral mass is $\sum_{\lambda<0}|\lambda|/\sum|\lambda|$. The final column reports the smallest eigenvalue of the mask itself.}
\label{tab:psd}
\end{table}
The table demonstrates directly that hard symmetric neighborhood masking can alter a structural property that the dense Gaussian kernel possesses exactly. Its purpose is finite-sample characterization at a size where the complete spectrum is available. PSD also has a natural diffusion interpretation. For a positive eigenvalue one may write $\lambda_m=e^{-t\nu_m}$ with $\nu_m\ge0$ and interpret $\nu_m$ as a real decay rate. Negative eigenvalues introduced by masking correspond to complex values of $\nu_m$ under this representation. Portions of the truncated spectrum can therefore lose the real diffusion-semigroup interpretation carried by the dense Gaussian kernel.

\subsection{Continuum geometry and ANN perturbations}\label{sec:knn:continuum}

Furthermore, $k$NN methods \citep{hein2007_graphLaplaciansRandomNeighborhoodGraphs,ting2010_convergenceGraphLaplacians, calder2022_spectralConvergenceGraphLaplacians}, using exact distances are still a $\mathcal{O}(N^2)$ theory. Approximate nearest-neighbor (ANN) methods introduce an additional perturbation. If $\widehat{A}_k$ is the returned approximate mask, then the effective kernel is:
\begin{align}
\widehat{\mathsf{K}}_{\beta,k}=\mathsf{K}_{\beta}\odot\widehat{A}_k.
\end{align}
Missed or substituted edges modify the weighted degrees, the $\alpha$-normalization, and ultimately the spectrum. Standard ANN recall \citep{malkov2018_hnsw} therefore gives only a partial measure of spectral accuracy because the effect of an incorrect edge depends on its Gaussian weight and on the spectral gaps of the modes being estimated \citep{boutsidis2009_cssp, drineas2008_cx}.
Some ANN implementations can additionally introduce run-to-run variability, depending on their construction and execution settings. The broader issue is the additional finite-sample perturbation introduced by approximate neighborhood construction, whose accuracy must be chosen before or adapted jointly with the bandwidth search. A neighborhood budget large enough to reproduce the dense-kernel support at the as-yet unknown $\beta^\star$ must either be selected conservatively in advance or increased as the bandwidth changes.

\subsection{Synthesis}\label{sec:knn:synthesis}

The comparison above separates several forms of approximation that affect diffusion-map calculations differently. A visually stable low-dimensional embedding can coexist with substantial changes in a larger spectral basis because the first few eigenvectors encode coarse geometry while higher eigenfunctions probe progressively shorter spatial scales. Matching the dense Gaussian kernel above a relative tolerance $\tau$ requires a neighborhood size scaling as:
\begin{align}
k^\ast(\beta,\tau)\sim N\left(\frac{\log(1/\tau)}{\beta}\right)^{d/2}.
\end{align}
At a bandwidth associated with $M$ resolved modes, this becomes:
\begin{align}
k^\ast\sim\frac{N\log(1/\tau)^{d/2}}{M}.
\end{align}
These relations characterize support fidelity of the dense finite-sample Gaussian operator. Spectral fidelity can persist under stronger truncation when the omitted interactions have limited influence on the target eigenspace, making direct eigenspace comparisons an additional empirical diagnostic.
The row-mass calculation provides a tolerance-independent measure of truncation strength. In the Swiss-roll experiment, $k=64$ retains $4.4\%$ of the Gaussian row mass at $\beta^\star$, while $k=1024$ retains approximately half. Hard neighborhood masking can also remove the PSD structure of the Gaussian kernel and alter the deeper portion of its diffusion spectrum.

These observations complement the main-text analysis of low-rank approximations. At bandwidths relevant to resolving many diffusion modes, the Gaussian matrix can simultaneously have substantial effective rank and substantial spatial support. Low-rank compression restricts the number of retained spectral directions, while aggressive fixed-$k$ sparsification removes a large fraction of spatial interactions. Both operations therefore modify the finite-sample Gaussian operator in distinct ways. 

%% file: sections/keops_compare.tex
\section{\textsc{KeOps} Comparison}\label{sec:keops}

Here we demonstrate a comparison of the \textsc{KeOps} library\footnote{Using \texttt{pip install pykeops==2.3} (Released: Apr 17, 2025)}, \cite{feydy2020_fastGeometricLearning, charlier2021_keops}, versus our matrix-vector (matvec) and matrix-matrix (matmat) primitive \textsc{XGEMM}. The results are shown in fig. \ref{fig:t6-kernel-race}, and both achieve similar timing results, and agree to \textit{machine-precision} (\texttt{fp32} agreement is within $\sim10^{-6}$).

\input{sections/keops}

%% file: sections/keops.tex

\pgfplotstableread[col sep=comma]{
N,keops,keops_std,xgemm_tilt,xgemm_tilt_std
1024,0.0004457973281,7.680200185e-05,0.001559583578,0.0003818804689
2048,0.0006339477708,2.115053364e-05,0.001320716083,0.0001524368059
4096,0.001037365833,1.863864618e-05,0.001380252333,0.00015299742
8192,0.001852067667,1.877979067e-05,0.001779931556,0.000130694434
16384,0.003450548,2.064877512e-05,0.00249223095,0.0001711784416
32768,0.006658661733,2.605012144e-05,0.0046829906,0.0001468164889
65536,0.02478053827,3.287704241e-05,0.010897011,0.0001407692908
131072,0.09745312275,5.3261146e-05,0.03901273713,9.496419481e-05
262144,0.3864002408,0.0002910280634,0.1407541988,0.001981973134
524288,1.420502707,0.02165109436,0.5305517392,0.0009434275468
1048576,5.533628987,0.0008291579239,2.08317415,0.01233083054
2097152,21.71997189,0.002487194098,8.825207709,0.002319376875
4194304,86.08497553,0.0002461417492,34.77380254,0.04694122922
8388608,342.7293724,0.009963519213,138.6485751,0.008448233222
16777216,1368.53809,0,552.5763896,0
}\tRaceMatvec

\pgfplotstableread[col sep=comma]{
N,keops,keops_std,xgemm_tilt,xgemm_tilt_std
1024,0.0003749358125,5.302090441e-05,0.001752765719,0.0005784329462
2048,0.000509945625,1.828895037e-05,0.001358330396,0.0001166604066
4096,0.0008075241944,2.081363468e-05,0.001390456528,0.0001401370191
8192,0.001402170148,2.313290507e-05,0.001734590704,0.0001545710489
16384,0.0025544201,1.978367183e-05,0.0022193506,0.00019719328
32768,0.0049185538,0.0001475460709,0.004509783533,0.0001612354645
65536,0.01263810309,1.916143741e-05,0.012966476,8.996735267e-05
131072,0.06330661775,2.891372864e-05,0.04318948075,0.0003355945499
262144,0.1968964182,0.0009703961665,0.152757719,0.002071955034
524288,0.7379114384,0.0009720442985,0.5869335358,0.0004408127247
1048576,2.944729413,0.0005129092096,2.897318638,0.0004728286718
2097152,11.54437859,0.002012968156,11.4596021,0.001145862733
4194304,45.83910731,0.0006576793101,45.80087754,0.003229941447
8388608,182.5580014,0.01074668452,182.570029,0.07421529166
16777216,731.2434026,0,731.0217599,0
}\tRaceMatmat

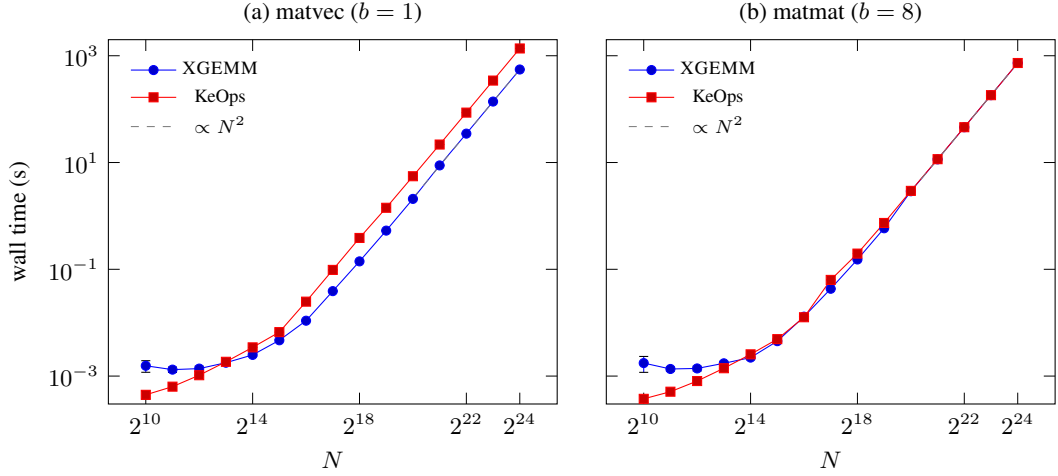
\begin{figure}[t]
\centering
\begin{tikzpicture}
\begin{groupplot}[
  group style={
    group size=2 by 1,
    horizontal sep=0.65cm,
    ylabels at=edge left,
    yticklabels at=edge left
  },
  scale only axis,
  width=0.425\textwidth,
  height=0.345\textwidth,
  xmode=log,
  log basis x=2,
  ymode=log,
  ymin=3e-4,
  ymax=2e3,
  xlabel={$N$},
  ylabel={wall time (s)},
  xtick={1024,16384,262144,4194304,16777216},
  xticklabels={$2^{10}$,$2^{14}$,$2^{18}$,$2^{22}$,$2^{24}$},
  tick label style={font=\small},
  label style={font=\small},
  legend style={font=\scriptsize, draw=none, fill=none},
  every axis title/.style={font=\small, at={(0.5,1.02)}, anchor=south},
]

\nextgroupplot[
  title={(a) matvec ($b=1$)},
  legend pos=north west,
]
\addplot+[mark=*, mark size=1.7pt,
  error bars/.cd, y dir=both, y explicit]
  table[x=N, y=xgemm_tilt, y error=xgemm_tilt_std]{\tRaceMatvec};
\addlegendentry{\textsc{XGEMM}}
\addplot+[mark=square*, mark size=1.7pt,
  error bars/.cd, y dir=both, y explicit]
  table[x=N, y=keops, y error=keops_std]{\tRaceMatvec};
\addlegendentry{KeOps}
\addplot[dashed, gray, domain=1048576:16777216, samples=2]
  {552.5763896*(x/16777216)^2};
\addlegendentry{$\propto N^2$}

\nextgroupplot[
  title={(b) matmat ($b=8$)},
  legend pos=north west,
]
\addplot+[mark=*, mark size=1.7pt,
  error bars/.cd, y dir=both, y explicit]
  table[x=N, y=xgemm_tilt, y error=xgemm_tilt_std]{\tRaceMatmat};
\addlegendentry{\textsc{XGEMM}}
\addplot+[mark=square*, mark size=1.7pt,
  error bars/.cd, y dir=both, y explicit]
  table[x=N, y=keops, y error=keops_std]{\tRaceMatmat};
\addlegendentry{KeOps}
\addplot[dashed, gray, domain=1048576:16777216, samples=2]
  {731.0217599*(x/16777216)^2};
\addlegendentry{$\propto N^2$}

\end{groupplot}
\end{tikzpicture}

\caption{Raw matrix-free Gaussian-kernel scaling on the centered anisotropic
flat torus $T^6$ ($D=32$, $\beta=1$) using one RTX--PRO--6000 GPU.
\textbf{(a)}~True matrix--vector application $Kv$ ($b=1$).
\textbf{(b)}~Eight-right-hand-side matrix--matrix application $KV$ ($b=8$).
\textsc{XGEMM}--TILT and KeOps evaluate the same dense Gaussian interaction
without materializing the $N\times N$ kernel matrix. Points show mean wall
time and error bars show one standard deviation over repeated timed
applications; the dashed guides indicate exact $\mathcal{O}(N^2)$ scaling in
the large-$N$ regime.}
\label{fig:t6-kernel-race}
\end{figure}